\documentclass[12pt]{article}

\usepackage{newtxtext,newtxmath}

\usepackage{graphicx}

\usepackage[letterpaper, textwidth=18cm, top=18mm, bottom=18mm]{geometry}

\PassOptionsToPackage{obeyspaces}{url}
\usepackage{xurl}
\usepackage{xr}
\usepackage{siunitx}
\usepackage{comment}
\usepackage{graphicx}%
\usepackage{multirow}%

\usepackage{amsmath,amssymb,amsfonts}%
\usepackage{amsthm}%
\usepackage{newtxmath}%
\usepackage[title]{appendix}%
\usepackage{xcolor}%
\usepackage{textcomp}%
\usepackage{manyfoot}%
\usepackage{booktabs}%
\usepackage{algorithm}%
\usepackage{algorithmicx}%
\usepackage{algpseudocode}%
\usepackage{listings}%
\usepackage[nolist,nohyperlinks]{acronym}
\usepackage{cleveref}
\usepackage{mathtools}
\usepackage{setspace}

\begin{acronym}
\acro{OC}{optimal control}
\acro{LQR}{linear quadratic regulator}
\acro{MAV}{micro aerial vehicle}
\acro{IMAV}{insect-scale micro aerial vehicle}
\acro{GPS}{guided policy search}
\acro{UAV}{unmanned aerial vehicle}
\acro{MPC}{model predictive control}
\acro{NMPC}{nonlinear model predictive control}
\acro{RTMPC}{robust tube nonlinear model predictive control}
\acro{BC}{behavior cloning}
\acro{DR}{domain randomization}
\acro{IL}{imitation learning}
\acro{DAgger}{dataset aggregation}
\acro{MDP}{Markov decision process}
\acro{CoM}{Center of Mass}
\acro{SD}{standard deviation}
\acro{MSE}{mean squared error}
\acro{RMSE}{root mean square error}
\acro{NN}{neural network}
\acro{UKF}{unscented Kalman filter}
\acro{PPO}{proximal policy optimization}
\acro{LfD}{learning from demonstration}
\acro{RL}{reinforcement learning}
\acro{PPO}{proximal policy optimization}
\acro{DEA}{dielectric elastomer actuator}
\acro{LSTM}{long short-term memory}
\acro{MoI}{moment of inertia}
\acro{IMU}{inertial measurement unit}
\acro{ToF}{time of flight}

\acro{AIO}{Airflow-Inertial Odometry}
\acro{BC}{Behavior Cloning}
\acro{CNN}{convolutional NN}
\acro{CoM}{center of mass}
\acro{DA}{data augmentation}
\acro{DAgger}{Dataset-Aggregation}
\acro{DDP}{differential dynamic programming}
\acro{DNN}{deep neural network}
\acro{DO}{disturbance observer}
\acro{DR}{Domain Randomization}
\acro{EKF}{extended Kalman filter}
\acro{GP}{Gaussian Process}
\acro{GPS}{Guided Policy Search}
\acro{IL}{Imitation Learning}
\acro{IMU}{Inertial Measurement Unit}
\acro{KKT}{Karush–Kuhn–Tucker}
\acro{LQR}{Linear Quadratic Regulator}
\acro{LSTM}{Long Short-Term Memory}
\acro{MAP}{Maximum a Posterior}
\acro{MAV}{Micro Aerial Vehicle}
\acro{MDP}{Markov Decision Process}
\acro{MLP}{Multi-Layer Perceptron}
\acro{MPC}{Model Predictive Control}
\acro{MRAC}{Model Reference Adaptive Control}
\acro{MRP}{Modified Rodriguez Parameters}
\acro{MSE}{Mean Squared Error}
\acro{NLP}{nonlinear program}
\acro{NN}{neural network}
\acro{NMPC}{Nonlinear Model Predictive Control}
\acro{NeRF}{Neural Radiance Field}
\acro{OC}{Optimal Control}
\acro{OCP}{Optimal Control Problem}
\acro{PF}{particle filter}
\acro{QP}{quadratic program}
\acro{RBF}{radial basis function}
\acro{RK}{Runge-Kutta}
\acro{RL}{Reinforcement Learning}
\acro{RMA}{Rapid Motor Adaptation}
\acro{RMSE}{Root Mean Squared Error}
\acro{RMS}{Root Mean Squared}
\acro{RNN}{recurrent neural network}
\acro{RTE}{Relative Translation error}
\acro{RTI}{real-time iteration}
\acro{SA}{Sampling Augmentation}
\acro{SAMA}{Sampling Augmentation for Motion Adaptation}
\acro{SD}{Standard Deviation}
\acro{SQP}{sequential quadratic program}
\acro{Tube-NeRF}{Tube-NeRF}
\acro{UAV}{Unmanned Aerial Vehicle}
\acro{UKF}{Unscented Kalman filter}
\acro{USQUE}{Unscented Quaternion estimator}
\acro{UT}{Unscented Transformation}
\acro{VIO}{Visual-Inertial Odometry}
\acro{VO}{visual odometry}
\acro{VSA}{Visuomotor Sampling Augmentation}
\acro{iLQR}{iterative linear quadratic regulator}
\acro{DAC}{Digital-to-Analog Converter}
\acro{UDP}{User Datagram Protocol}

\end{acronym}

\newif\ifCameraReady

\CameraReadytrue 

\ifCameraReady

\else

    \usepackage[left]{lineno}
    \linenumbers
\fi

\renewenvironment{abstract}
	{\quotation}
	{\endquotation}

\date{}

\makeatletter
\renewcommand{\fnum@figure}{\textbf{Figure \thefigure}}
\renewcommand{\fnum@table}{\textbf{Table \thetable}}
\makeatother

\usepackage{scicite}

\usepackage{url}

\def\scititle{
	Controlled Flight of an Insect-Scale Flapping-Wing Robot via Integrated Onboard Sensing and Computation 
}
\title{ \Large \textbf{\scititle}}

\author{
	\normalsize Yi-Hsuan Hsiao$^{1\dagger}$,
	Quang Phuc Kieu$^{1\dagger}$,
	Zhongtao Guan$^{1\dagger}$,
    Suhan Kim$^{1}$, 
    Jiaze Cai$^{1}$, \and
    \normalsize Owen Matteson$^{2}$,
    Jonathan P. How$^{2\ast}$,
    Elizabeth Farrell Helbling$^{3\ast}$,
    YuFeng Chen$^{1\ast}$ \\
    \color{white}\tiny.\\
    \footnotesize$^{1}$Department of Electrical Engineering and Computer Science, Massachusetts Institute of Technology, \\
    \footnotesize Cambridge, Massachusetts, 02139, United States.\\
    \footnotesize$^{2}$Department of Aeronautics and Astronautics, Massachusetts Institute of Technology, \\
    \footnotesize Cambridge, Massachusetts, 02139, United States.\\
    \footnotesize$^{3}$School of Electrical and Computer Engineering, Cornell University,\\
    \footnotesize Ithaca, New York, 14850, United States.\\
	\footnotesize$^\ast$Corresponding authors. Emails: jhow@mit.edu, farrell@cornell.edu, yufengc@mit.edu \\
	\footnotesize$^\dagger$These authors contributed equally to this work.
}

\begin{document}

\linespread{0.9}\selectfont
\maketitle
\linespread{1.5}\selectfont

\begin{abstract} 
\begingroup
\setstretch{1.2}  
\bfseries \boldmath

Aerial insects can effortlessly navigate dense vegetation, whereas similarly sized aerial robots typically depend on offboard sensors and computation to maintain stable flight. This disparity restricts insect-scale robots to operation within motion capture environments, substantially limiting their applicability to tasks such as search-and-rescue and precision agriculture. In this work, we present a 1.29-gram aerial robot capable of hovering and tracking trajectories with solely onboard sensing and computation. The combination of a sensor suite, estimators, and a low-level controller achieved centimeter-scale positional flight accuracy. Additionally, we developed a hierarchical controller in which a human operator provides high-level commands to direct the robot’s motion. In a 30-second flight experiment conducted outside a motion capture system, the robot avoided obstacles and ultimately landed on a sunflower. This level of sensing and computational autonomy represents a significant advancement for the aerial microrobotics community, further opening opportunities to explore onboard planning and power autonomy. 
\endgroup

\end{abstract}

\paragraph*{Short Title:} Sensory autonomy in a micro-aerial-robot

\paragraph*{One-Sentence Summary:} A 1.29-g flapping-wing robot performs controlled flights with onboard sensors and computation.  

\newpage

\paragraph*{Main Text:}
\subsection*{INTRODUCTION}\label{sec:intro}

Aerial insects rely on a complex, hierarchical sensory system to perform feedback-controlled flight. To maintain low-level flight stability, fruit flies integrate inputs from halteres \cite{pringle1948gyroscopic}, antennae \cite{kamikouchi2009neural}, and ocelli \cite{krapp2009ocelli} to estimate rotational rates, gravitational orientation, and the visual horizon. At a higher level of control, compound eyes \cite{srinivasan2011visual} and mechanoreceptors \cite{tuthill2016mechanosensation} provide visual and haptic cues that support navigation, landing, and takeoff. This sensory information is processed by a compact nervous system comprising approximately 140,000 neurons and 50 million synapses \cite{reardon2024largest}. This highly efficient sensing and control architecture, combined with the small size of aerial insects, enables capabilities surpassing that of existing aerial robots. For example, in response to the recent Myanmar earthquake, cyborg cockroaches equipped with infrared cameras were deployed in search-and-rescue operations. These systems successfully navigated debris-filled environments and searched for survivors, demonstrating the potential of small-scale bio-inspired platforms for disaster relief applications.

Inspired by aerial insects, researchers in microrobotics have developed numerous insect-scale micro aerial vehicles (IMAVs) \cite{kim2025acrobatics,ma2013controlled,bena2023high,chukewad2021robofly,xiang2025high,drew2018toward,ozaki2021wireless} and achieved exciting progress. These platforms have demonstrated insect-like flight behaviors, including hovering \cite{ma2013controlled}, body saccades \cite{hsiao2025aerobatic}, somersaults \cite{kim2025acrobatics}, collision recovery \cite{chen2021collision}, perching \cite{graule2016perching}, landing \cite{hyun2025sticking}, and multimodal locomotion \cite{chen2017biologically}. Despite these impressive capabilities, achieving fully autonomous flight remains a major challenge for IMAVs due to strict payload limitations, high power consumption, and rapid body dynamics. Consequently, most IMAVs are restricted to small flight volumes ($<$ 50 cm × 50 cm × 50 cm) as they rely on offboard sensing, computation, and energy sources. To address power autonomy, prior work has developed custom boost electronics and lightweight solar panels, enabling short-duration flights powered by laser illumination \cite{james2018liftoff} or high-intensity light sources \cite{jafferis2019untethered}. However, on the sensory side, no IMAV to date has demonstrated flight relying exclusively on onboard sensors and a microcontroller unit (MCU). Currently, the 18-g quadcopter Funfliber-Drone \cite{muller2021funfiiber} represents the smallest aerial robot capable of fully autonomous sensing and computation, yet it remains an order of magnitude heavier than typical IMAVs. The gap on sensory and compute autonomy represents one of the most critical challenges facing the microrobotics community.

To address the stringent constraints on size, speed, weight, and power, researchers have proposed and investigated a variety of sensing solutions over the past decade. Early studies focused on the design and characterization of individual sensors, including ocelli \cite{fuller2014controlling}, magnetometer \cite{helbling2014pitch}, proximity sensor \cite{helbling2018altitude}, and inertial measurement unit (IMU) \cite{fuller2014using}. More recent efforts have integrated multiple sensors \cite{yu2025tinysense,talwekar2022towards} and evaluated their performance on larger-scale quadcopters or robotic arms \cite{naveen2025hardware}. Researchers have also developed altitude and attitude estimation algorithms deployable on small MCUs \cite{yu2025tinysense}. Despite optimization of individual design metrics—such as sensor mass, measurement drift, and computational cost—none of these sensor suites can be deployed on IMAVs to enable autonomous flight. To address this limitation, we propose a holistic design approach that jointly considers robot payload capacity, power budget, and target flight performance. Specifically, our sensor suite and MCU design satisfy four high-level criteria: (1) the vehicle maintains precise maneuverability after sensor integration; (2) sensing and computation power consumption remains substantially lower than that of the flight actuators; (3) onboard altitude and attitude measurements achieve sufficient accuracy to support higher-level tasks such as setpoint tracking, obstacle avoidance, and pollination; and (4) the design is extensible, allowing integration of additional sensors to enable new functionalities.

In this work, we present a 1.29-g insect-scale flapping-wing robot capable of performing flight with onboard sensing and computation. The 244-mg flight package integrates an IMU, a time-of-flight (ToF) sensor, an optical flow sensor, and a dual-core MCU. Using these sensors, we developed a state estimator that computes the full six-degrees-of-freedom robot state with high accuracy (errors around 3.05$^\circ$ and  2.5 cm) and low drift (around 1.2 $^\circ$/s and 0.5 cm/s). The robot successfully executed a sequence of sensory autonomous hovering and trajectory-tracking flights with 3.96-cm root-mean-square (RMS) positional error. Low-level flight stability and accuracy further enabled higher-level control capabilities. We implemented a hierarchical control architecture that allows a human operator to issue real-time trajectory commands. Under high-level position commands, our robot could fly outside of a motion capture system, evade obstacles, and precisely land on a sunflower. The estimator and controller execute on a single MCU core, leaving additional computational resources available for future onboard implementation of communication, image processing, and motion planning. This integrated design methodology addresses a long-standing challenge in IMAV research and opens new opportunities for pursuing power-autonomous flight.

\begin{figure}[ht]
\centering
\includegraphics[width=180mm]{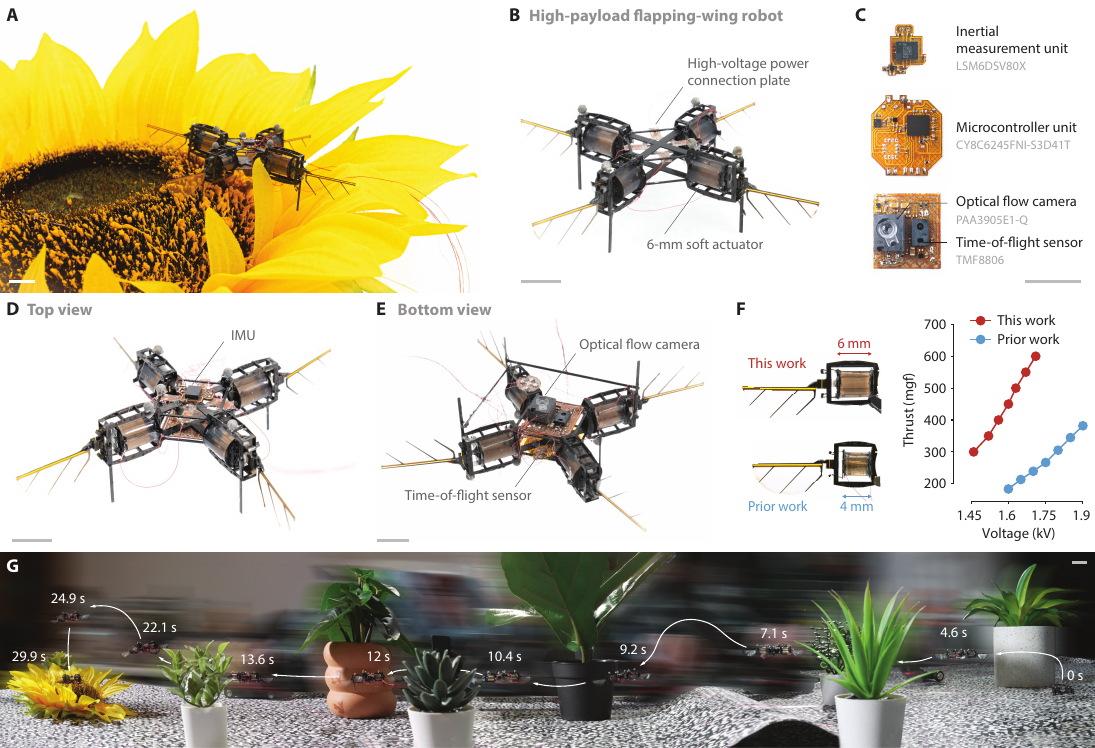}
\vspace{-8mm}
\caption{\textbf{Overview of robot design, flight package selection, and sensory autonomous flight.} A) An image illustrating the robot lands on a sunflower. B) An image of the robot without integrating sensors. C) The three PCB boards carry an IMU (top), an MCU (center), and an optical flow and a ToF (bottom), respectively. The model numbers are reported in grey text. D-E) Top (D) and bottom (E) views of the 1.29-g robot with the integrated sensors and MCU. F) Lift force comparison shows the scaled-up robot module (red) achieves 46\% lift increase compared to a prior design (blue). G) A side view composite image showing a flight performed outside a motion capture arena. The robot avoids obstacles and lands on a sunflower. The background is blurred to remove facial information of the experimenter. Scale bars in (A-E) and (G) represent 1 cm. }\label{fig:overview}
\end{figure}

\subsection*{RESULTS}\label{sec:results}

\subsubsection*{Task driven robot design and flight avionics selection}

We adopted an integrated design approach guided by the intended flight tasks. In particular, we envisioned the application of IMAVs in assisted pollination scenarios, where they must navigate cluttered environments, avoid obstacles, and interact safely with delicate flowers. To enable perching or landing on a flower (Fig. \ref{fig:overview}A), the robot must be lightweight (approximately 1 g) and robust to in-flight collisions. Furthermore, effective avoidance of stationary obstacles requires a high-bandwidth low-level controller ($>$100 Hz) capable of maintaining near-centimeter positional accuracy, while operating under commands from a comparatively slow high-level planner ($\sim$1 Hz).

Achieving these performance objectives was challenging because IMAVs face three distinct constraints compared to mesoscale ($>$10 g) aerial robots: (1) rapid body dynamics that necessitate high-rate sensing and control; (2) high-frequency body oscillations that can amplify sensor drift; and (3) severe limitations in payload and power budget. To address these challenges, we defined three quantitative design criteria. First, based on prior flight experiments \cite{hsiao2025aerobatic}, we specified ideal minimum update rates of 200 Hz for attitude sensing and 200 Hz for the control loop. To achieve sustained stable flight ($>$10 s) with low positional drift ($<$5 cm), the state estimation accuracy was required to be within 4 cm for position and 10° for attitude over a 10-s horizon. Second, previous flight studies \cite{chen2019controlled,chen2021collision} indicated that a minimum lift-to-weight ratio of 1.3 is necessary for accurate trajectory tracking, which imposed constraints on vehicle scaling and net payload. Third, to support future investigations of power-autonomous flight, we required that sensing and onboard computation consume less than 10\% of the total power budget.

These criteria guided robot redesign and sensor selection. Our robot (Fig. \ref{fig:overview}B-E) was comprised of four independent flapping-wing modules, each actuated by a dielectric elastomer actuator (DEA) at a flapping frequency of 330 Hz. Compared to prior work \cite{kim2025acrobatics}, the DEA length and the transmission ratio were scaled up by 1.5 times, and the net lift force of each module was increased from 410 mg \cite{kim2025acrobatics} to 600 mg (Fig. \ref{fig:overview}F). The new robot weighed 1.05 g excluding the sensor package, and it achieved a 1.9 lift-to-weight ratio after carrying a 244-mg payload. Under hovering conditions, the artificial flight muscles consumed 2.04 W of power.

To achieve high-bandwidth and accurate state estimation, we developed a flight electronics package comprising a 14-mg IMU (LSM6DSV80X), a 13-mg ToF (TMF8806), a 90-mg optical flow sensor (PAA3905E1-Q), and a 7-mg MCU (CY8C6245FNI-S3D41T). These components were integrated onto three custom-designed flexible printed circuit boards (PCBs) (Fig. \ref{fig:overview}C) and mounted within the robot’s central payload compartment (Figs. \ref{fig:overview}D-E). To ensure accurate measurement of rotational dynamics, the IMU was positioned on the top PCB near the robot’s geometric center (Fig. \ref{fig:overview}D). Because the IMU is sensitive to mechanical vibrations and thermal fluctuations, we used a 0.4-mm thick fiberglass substrate to mitigate these effects. The ToF and optical flow sensors were oriented downward and mounted on the bottom PCB to measure altitude and lateral velocity relative to ground. Most flight experiments were conducted using configurations that included only the top and bottom sensor boards. Fig. \ref{fig:overview}G shows a sensory autonomous flight where the robot takes off, evades obstacles, and ultimately lands on a sunflower.

To further demonstrate onboard computation, we designed an MCU board that can be installed between the two sensor boards (Fig. \ref{fig:overview}C). In this configuration, all sensors were directly interfaced with the MCU via I²C or other serial communication protocols. The complete flight package weighed 244 mg, with the sensor boards and the MCU board accounting for 15\% and 4\% of the total mass, respectively. Under the highest sensing rates and computational loads, the flight electronics consumed 120 mW, corresponding to 6\% of the power consumed by the flight actuators. Additional details on PCB fabrication, component mass, power distribution, and sensor characterization are provided in Supplementary Text.

\subsubsection*{Estimation strategy}
\begin{figure}[ht]
\centering
\includegraphics[width=127mm]{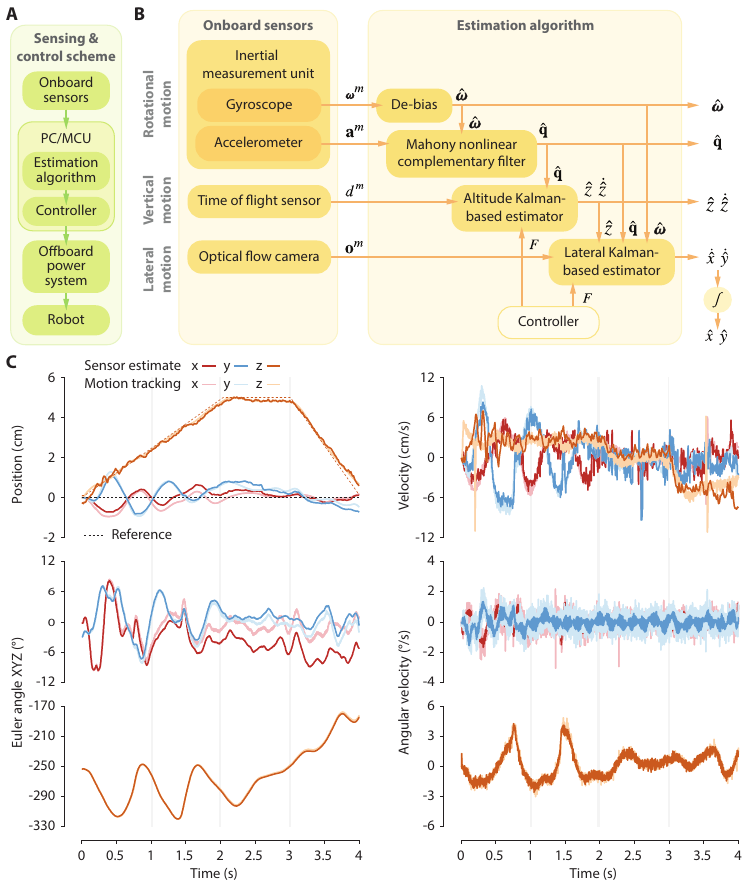}
\vspace{-4mm}
\caption{\textbf{Onboard sensing and state estimation.} A) A high-level schematic of the sensing, estimation, and control pipeline. B) A schematic of the sensing and cascaded estimator design. A complementary filter and two observers sequentially estimate robot attitude, altitude, and lateral position, respectively. C) Evaluation of state estimation accuracy. The robot conducts a 4-s flight under feedback control with offboard motion capture data. The onboard measured and estimated position, velocity, attitude, and angular velocity are compared against the offboard motion capture data. }\label{fig:observer}
\end{figure}

Performing flight requires five major elements: sensors, estimators, controller, power generation, and robotic hardware (Fig. \ref{fig:observer}A). Based on sensor measurements and robot dynamics, we designed a strategy to calculate the robot orientation $\mathbf{q}$, angular velocity $\boldsymbol{\omega}$, position $[x, y, z]$, and translational velocity $[\dot{x}, \dot{y}, \dot{z}]$. The cascaded framework (Fig. \ref{fig:observer}B) comprises three major components that estimate attitude ($\hat{\mathbf{q}}$, $\hat{\boldsymbol{\omega}}$), altitude ($\hat{z}$, $\dot{\hat{z}}$), and lateral motion ($\dot{\hat{x}}$,$\dot{\hat{y}}$).

First, based on IMU measurements, we implemented the Mahony nonlinear complementary filter \cite{mahony2008nonlinear} to estimate the attitude states $\hat{\mathbf{q}}$  and $\hat{\boldsymbol{\omega}}$, where $\hat{\mathbf{q}}$ and $\hat{\boldsymbol{\omega}}$ are the orientation represented by quaternion and the angular velocity with respect to the body reference frame, respectively. The gyroscope and accelerometer inside the IMU measured angular velocity $\boldsymbol{\omega}^m$ and linear acceleration $\mathbf{a}^m$ with a sample rate of 480 Hz. $\hat{\boldsymbol{\omega}}$  and $\hat{\mathbf{q}}$ are given by: 
\begin{align}
\hat{\boldsymbol{\omega}} = \boldsymbol{\omega}^m - \hat{\mathbf{b}}_{\omega},\\
\hat{\mathbf{q}} = f_{\text{Mah}}(\mathbf{a}^m, \hat{\boldsymbol{\omega}}),
\end{align}
where $\hat{\mathbf{b}}_{\omega}$ is the estimated gyroscope bias based on prior flights. 

Second, based on attitude estimates, distance measured by the ToF sensor, and robot dynamics, we designed a Kalman filter to estimate the altitude states:
\begin{align}
\begin{bmatrix}
\hat{z} \\
\dot{\hat{z}}
\end{bmatrix}
= f_{\text{KF,alt}}(\hat{\mathbf{q}}, d^m, F).
\end{align}
Here, $\hat{z}$ and $\dot{\hat{z}}$ represent the estimated altitude and vertical velocity, $d^m$ is the distance measurement from the \ac{ToF} sensor; and $F$ is the thrust command from the controller.

Lastly, we designed the third estimator that calculated robot lateral velocity and position based on attitude and altitude estimates, optical flow measurements, and the robot dynamics. The velocity estimate is computed from another Kalman filter: 
\begin{align}
\begin{bmatrix}
\dot{\hat{x}}\\
\dot{\hat{y}}
\end{bmatrix}
= f_{\text{KF,lat}}(\hat{\mathbf{q}}, \hat{\boldsymbol{\omega}}, \hat{z}, \mathbf{o}^m, F).
\end{align}
Here, $\dot{\hat{x}}$ and $\dot{\hat{y}}$ are the estimated lateral velocities, and $\mathbf{o}^m$ denotes the pixel measurement vector from the optical flow camera. The lateral positions are obtained by integrating the velocity estimates:
\begin{align}
\hat{x}(t) = \int_{0}^{t} \dot{\hat{x}}(\tau) \, d\tau, \;\; \hat{y}(t) = \int_{0}^{t} \dot{\hat{y}}(\tau)\, d\tau.
\end{align}
Details of the Mahony filter and the altitude and position Kalman filters are described in Materials and Methods. 

Unlike other work that formulated an extended Kalman filter to estimate attitude and altitude within a single loop \cite{talwekar2022towards}, our design substantially reduces the computational cost because it does not require linearization and Jacobian calculation. Our Kalman filter implementation merely multiplies two-by-two matrices – much simpler than estimators where all states are coupled \cite{talwekar2022towards,naveen2025hardware}. Furthermore, the implementation of three distinct estimators addresses challenges arising from different sensor sampling rates. To evaluate this design, we conducted flight experiments and compared onboard estimation with motion capture data (Fig. \ref{fig:observer}C). The RMS errors of estimated attitude, body angular velocity, position, and translational velocity were 1.8 °, 0.8 °/s, 0.2 cm, and 2.6 cm/s, respectively. In this experiment, the feedback controller relied on motion capture data (Vicon Vantage V5 cameras), while onboard sensing and state estimation were recorded for post analysis. The following section presents flight results based exclusively on onboard sensing.

\subsubsection*{Flights with onboard sensing and offboard computation}
We evaluated the performance of the sensor suite and state estimator through five types of flight experiments. In this section, all flights were conducted using onboard sensing but offboard computation. The offboard estimator and controller were executed on a desktop computer using MathWorks Simulink, which also recorded flight data, including robot states and intermediate variables. For validation, the robot’s motion was independently measured using an external motion capture system that served as ground truth.

\begin{figure}[!ht]
\centering
\includegraphics[width=180mm]{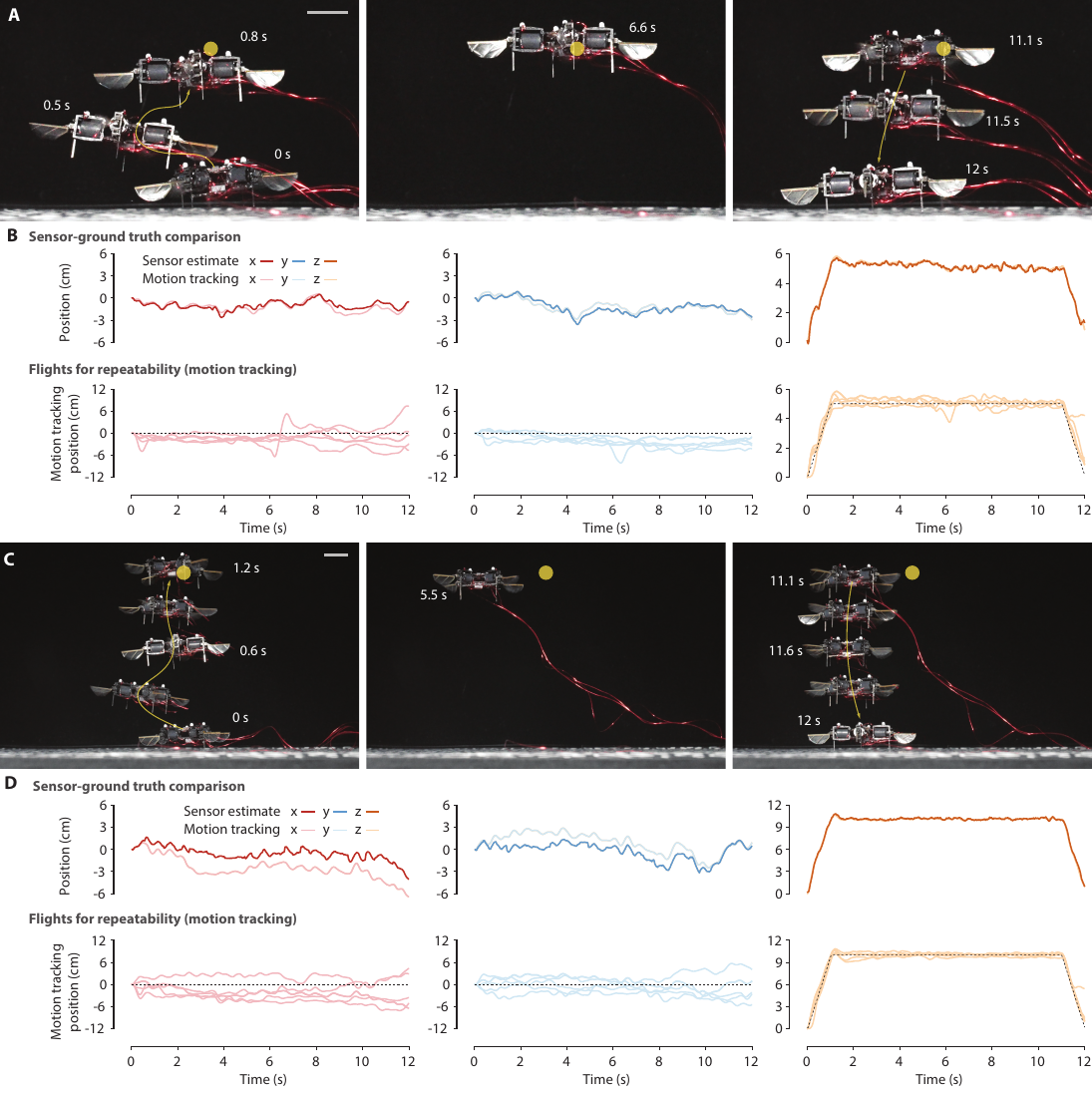}
\vspace{-8mm}
\caption{\textbf{Hovering demonstration with onboard sensing.} A) A composite image sequence of a 12-s hovering flight that is 5 cm above ground. B) Comparison of position measurements based on either onboard estimation or offboard motion capture. C) Measured position data of five repeated experiments. In (B-C), the three columns correspond to the x, y, and z positions, respectively.  D-F) Composite image sequence, position comparison, and repeated experiments of 12-s hovering flights set 10 cm above ground. Yellow dots in (A) and (C) represent the hovering setpoint. Scale bars in (A) and (C) represent 1 cm.}\label{fig:hovering}
\end{figure}

First, the robot performed a 12-s hovering flight at a commanded altitude of 5 cm above the ground (Fig. \ref{fig:hovering}A). By comparing the commanded setpoint with ground truth measurements, the RMS errors in lateral position and altitude were found to be 3.24 cm and 0.25 cm, respectively (Fig. \ref{fig:hovering}B). These tracking errors arise from a combination of state-estimation inaccuracies and flight-control performance. A comparison between the estimated states derived from onboard sensors and the measured states obtained from the motion-capture system is shown in Fig. \ref{fig:hovering}B. The RMS errors attributable to state estimation alone were 1.8 cm in lateral position and 0.1 cm in altitude. Over the 12-s flight duration, a mean lateral position drift of 2.0 cm was observed, which can be attributed to the integration of lateral velocity estimates from the optical flow sensor. This drift is expected to increase with flight altitude due to reduced pixel resolution.

To quantify this effect in our robot, we conducted an additional 12-s hovering flight at a commanded altitude of 10 cm above the ground (Fig. \ref{fig:hovering}C). At this height, the lateral position and altitude RMS errors were 3.65 cm and 0.26 cm, respectively (Fig. \ref{fig:hovering}D). We observed a 12.7\% increase of lateral position error at a higher altitude, showing the accuracy of optical flow estimate is height-dependent. In contrast, the altitude error remained small because the ToF sensor directly measures distance. Both of these hovering demonstrations were repeated five times to assess system repeatability and reliability (Fig. \ref{fig:hovering}B,D). Compared to the hovering flight performed under a motion capture system (Fig. \ref{fig:observer}C), these results illustrate that onboard sensing can achieve comparable flight accuracy at the centimeter level.

\begin{figure}[!ht]
\centering
\includegraphics[width=180mm]{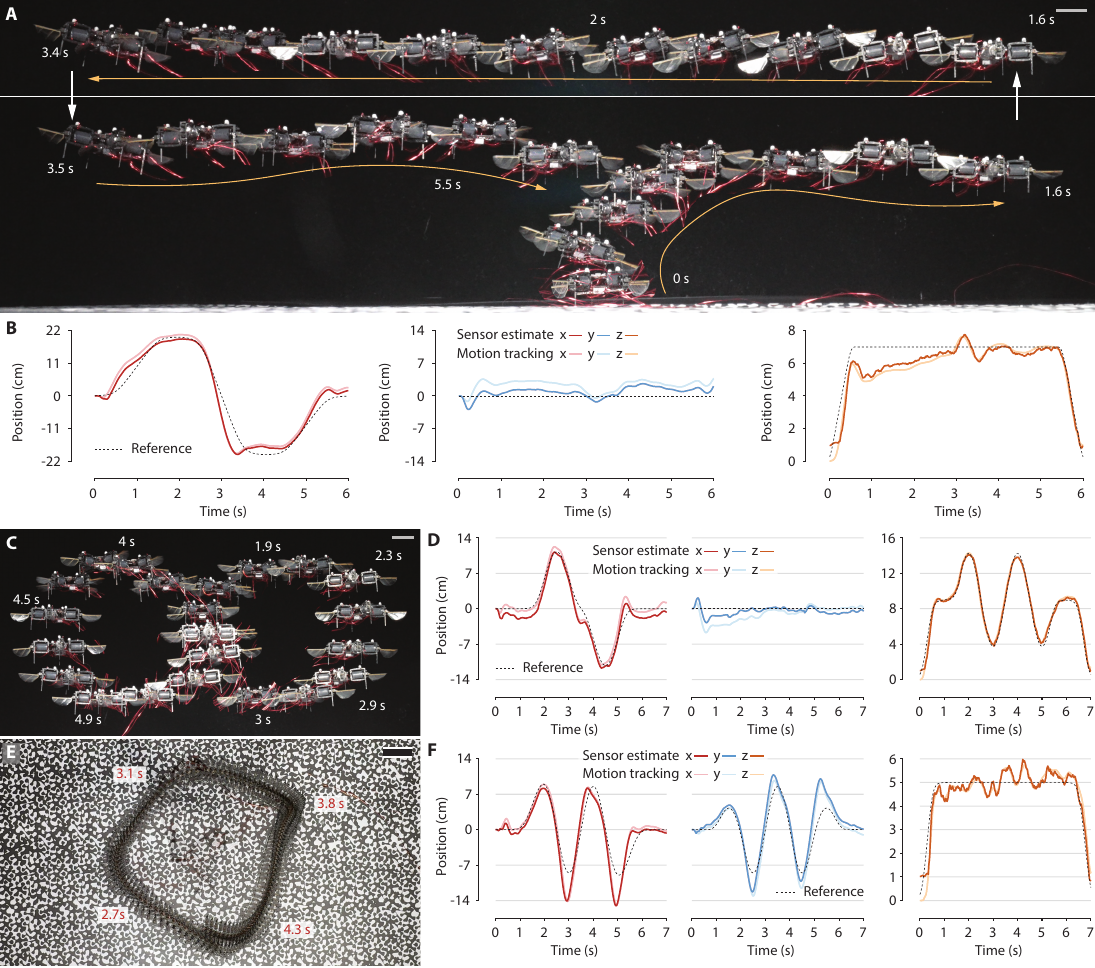}
\vspace{-8mm}
\caption{\textbf{Trajectory tracking demonstrations with onboard sensing.} A) A side-view composite image sequence of a 6-s setpoint switching flight. B)  Desired, estimated, and measured robot position corresponding to the experiment in (A). C) A side-view composite image in which the robot tracks an “infinity” sign. D) Desired, estimated, and measured robot position corresponding to the experiment in (C). E) A top-view composite image in which the robot tracks a planar circle. F) Desired, estimated, and measured robot position corresponding to the experiment in (E). Scale bars in (A), (C), and (E) represent 1 cm.}\label{fig:trajectory}
\end{figure}

Next, we conducted three types of trajectory-tracking flights. In the setpoint-tracking experiment, the robot flew between two locations separated by 40 cm (Fig. \ref{fig:trajectory}A). During the 7-s flight, it reached a maximum velocity of 68.4 cm/s, with RMS errors of 0.78 cm in altitude and 4.52 cm in lateral position. Fig. \ref{fig:trajectory}B compares the desired, estimated, and motion-capture trajectories, indicating that the observed position errors were dominated by flight-controller errors rather than state-estimation errors. The position state-estimation errors remained within 1.96 cm relative to the motion-capture measurements. 

We further evaluated tracking performance for smooth trajectories on both vertical and horizontal planes (Figs. \ref{fig:trajectory}C and \ref{fig:trajectory}E). When tracking a 22 cm × 10 cm figure-eight trajectory over 7 s, the RMS altitude and position errors were 0.29 cm and 2.19 cm, respectively (Fig. \ref{fig:trajectory}D). The robot also successfully tracked an 18 cm × 18 cm planar circular trajectory that was 5 cm above ground, and it achieved a maximum speed of 59.3 cm/s during the 7-s flight. In this case, the RMS altitude and position errors increased to 0.34 cm and 6.09 cm, respectively, due to the combined effects of yaw rotation and coupled motion along the x and y axes (Fig. \ref{fig:trajectory}F).

To assess flight repeatability, all three trajectory-tracking experiments were repeated three times. Across these dynamic maneuvers, state-estimation errors increased as the desired attitude changed more rapidly, indicating that integration errors in attitude estimation can propagate into position estimation. Nevertheless, these results demonstrate that the proposed low-level sensing and estimation framework provides sufficient flight precision over a 10-s horizon. Further reduction of sensor drift would require the incorporation of more sensors and a higher-level localization and planning algorithm.

\subsubsection*{Obstacle avoidance and precision landing under joystick control}
\begin{figure}[!htb]
\centering
\includegraphics[width=180mm]{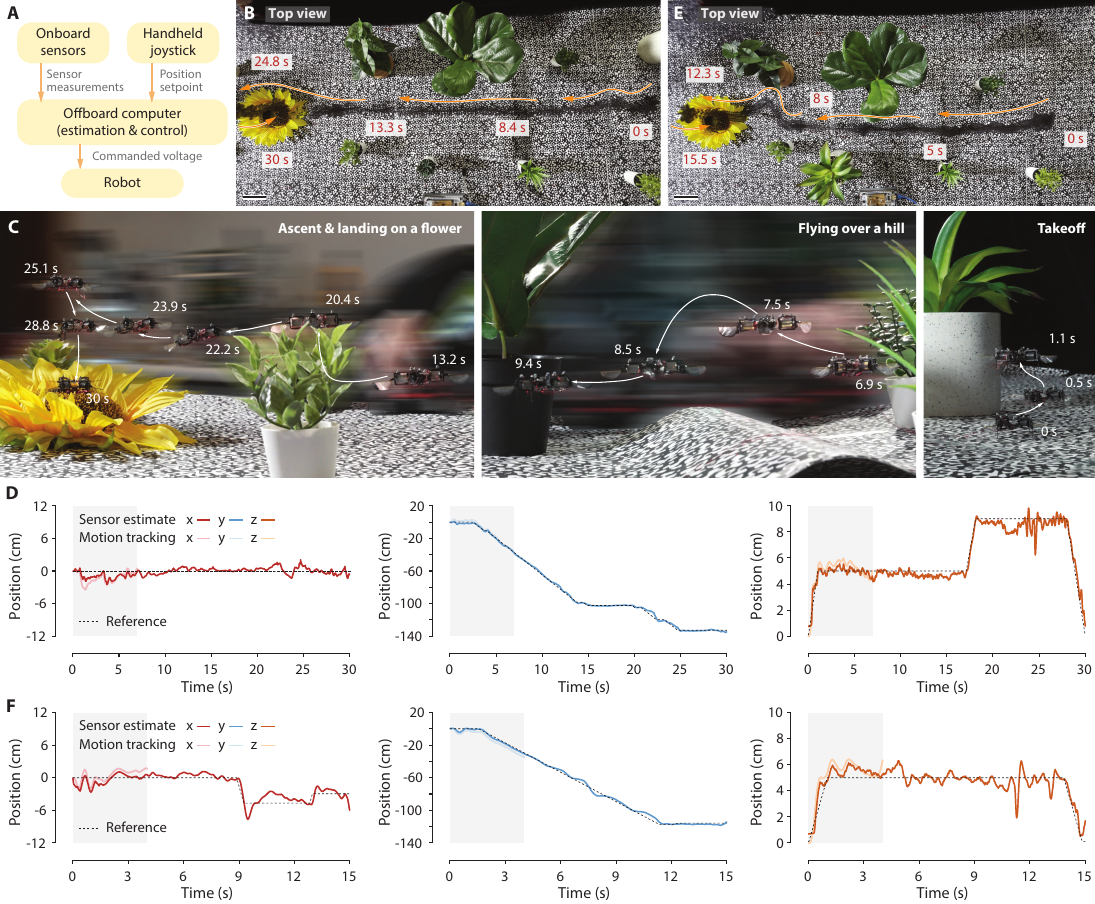}
\vspace{-8mm}
\caption{\textbf{Pollination demonstration.} A) A schematic of a hierarchical controller that integrates high-level human issued positional commands and low-level position tracking with onboard sensing. B) A top-view composite image sequence of a 30-s flight demonstration. C) Side-view image sequence that illustrates robot takeoff, obstacle traversal, flight altitude control, and landing. D) Desired, estimated, and measured robot positions corresponding to (B-C). E) A top-view composite image sequence of a 15-s flight demonstration. F) Desired,estimated, and measured robot positions corresponding to (E). In (D) and (F), the grey regions represent portion of the flights where motion tracking is available. Motion capture became unavailable when the robot flew out of the capture flight volume.  }\label{fig:operator}
\end{figure}
The proposed low-level sensing, state estimation, and control framework can be integrated with a high-level planner to enable operation in realistic, unstructured environments. To demonstrate this capability, we designed a hierarchical control architecture in which a human operator issued high-level commands via joystick (Fig. \ref{fig:operator}A). Specifically, the desired flight altitude and position were updated incrementally in response to joystick input. The robot was deployed in an environment populated with plants and a sunflower. After takeoff from the ground, an operator guided the robot around obstacles and ultimately executed a landing maneuver on the sunflower. To emphasize that the robot relied exclusively on onboard sensing, the obstacles and sunflower were placed outside of the motion capture arena. 

In the first flight demonstration (Fig. \ref{fig:operator}B and Movie S4), the operator cautiously guided the robot along a 1.2-m trajectory over a duration of 30 s. The flight sequence comprised several phases, including takeoff, traversal of a curved obstacle, a straight flight segment, ascent, and landing (Fig. \ref{fig:operator}C). Following liftoff, the robot autonomously cleared the curved obstacle without requiring additional high-level commands, demonstrating that the ToF sensor provided sufficient accuracy for obstacle detection and altitude regulation. The robot then proceeded toward the sunflower under real-time operator supervision. Prior to reaching the flower, the operator increased the commanded flight altitude to mitigate aerodynamic ground effects and reduce potential optical-flow estimation errors induced by the sunflower’s surface. Finally, the robot executed a controlled descent and successfully landed on the flower. A comparison between commanded and estimated robot position is shown in Fig. \ref{fig:operator}D. This demonstration illustrates that the robot can reliably execute high-level commands and achieve centimeter-scale landing accuracy in a cluttered environment.    

We conducted a second flight experiment in which obstacles were positioned closer together. The operator guided the robot along a path of comparable length in 15 s, issuing turning commands to navigate around the obstacles (Fig. \ref{fig:operator}E and Movie S4). In this trial, the ToF sensor autonomously detected the height change due to the flower, allowing the low-level controller to adjust flight altitude without operator commands. This flight was approximately twice as fast as the first demonstration, and the estimated trajectories from both flights are compared in Fig. \ref{fig:operator}F.

A third flight outside of the motion-capture arena was performed to evaluate robot reliability. These experiments demonstrate that the robot can successfully operate in realistic, cluttered environments using only onboard sensing, highlighting its potential for insect-like tasks such as pollination or navigation through confined spaces. In these demonstrations, computation was performed offboard to accommodate high-level operator commands and to record sensor and estimator data. In the following section, we present experiments integrating both onboard sensing and computation.

\subsubsection*{Flights with onboard sensing and computation}
\begin{figure}[ht]
\centering
\includegraphics[width=180mm]{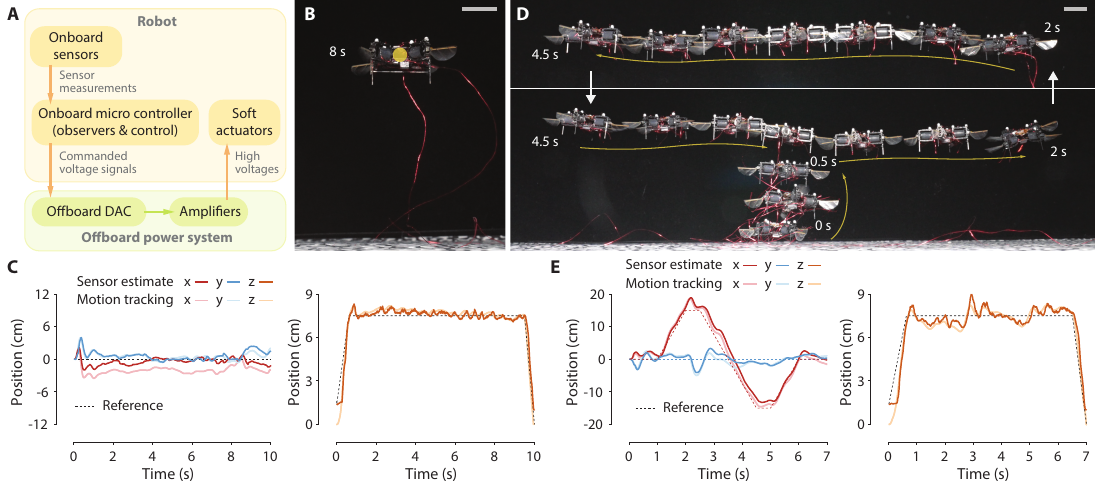}
\vspace{-8mm}
\caption{\textbf{Sensory and compute autonomous flight demonstrations.} A) A schematic that illustrates sensing and computation are done onboard while driving voltage is generated offboard. B) An image that shows a 10-s hovering flight. The yellow dot indicates the hovering setpoint. C) Estimated and measured robot position corresponding to the flight in (B). D) A composite image sequence of a 7-s setpoint switching flight. E) Estimated and measured robot position corresponding to the flight in (D). The scale bars in (B) and (D) represent 1 cm.}\label{fig:mcu}
\end{figure}
To evaluate flight performance using onboard sensing and computation, the MCU board was installed on the robot. All sensors were directly interfaced with the MCU, eliminating four wired connections to offboard computers at the cost of an additional 53.6 mg of payload. The estimator and controller were compiled and deployed on the MCU using MathWorks Embedded Coder. To reduce computational load and memory usage, the controller was converted from double-precision to single-precision floating-point format. The controller feedback rate was set to 480 Hz, and all computations were executed on a single MCU core. At the same data-logging rate, the MCU memory could store raw sensor measurements, estimated states, and actuator outputs for 15.6 s. Upon completion of each flight, the stored data were transferred to an offboard computer. The sensing and control architecture is shown in Fig. \ref{fig:mcu}A, and further details regarding the MCU controller implementation are provided in Supplementary Text.

Equipped with the onboard sensors and the MCU, the robot performed two types of flights: a 10-s hovering maneuver and a 7-s setpoint-switching trajectory (Figs. \ref{fig:mcu}B-D and Movie S5). To evaluate flight performance with onboard sensors and MCU, we recorded the flight data with a motion-capture system for post-experiment analysis. In the first flight, the robot hovered at 7.5 cm above the ground for 10 s. Figure \ref{fig:mcu}C shows that the onboard and offboard state measurements and controller outputs closely matched each other. Based on motion-capture data, the RMS errors in altitude and lateral position were 0.32 cm and 2.17 cm, respectively. In the second flight, the robot tracked two setpoints separated by 30 cm over 7 s (Fig. \ref{fig:mcu}D), achieving a maximum lateral velocity of 19.8 cm/s. The RMS errors in altitude and lateral position were 0.57 cm and 2.21 cm, respectively. Figure \ref{fig:mcu}E compares the onboard and offboard state measurements and the reference trajectory. These results demonstrate that onboard sensing and computation achieve similar accuracy compared to configurations using onboard sensing with offboard computation (Fig. \ref{fig:trajectory}B). All low-level sensor communication, state estimation, and control updates were executed on a single MCU core, leaving the computational capacity of the second core available for future implementation of high-level motion planning.
 
\newpage
\subsection*{DISCUSSION}
\begin{figure}[ht]
\centering
\includegraphics[width=180mm]{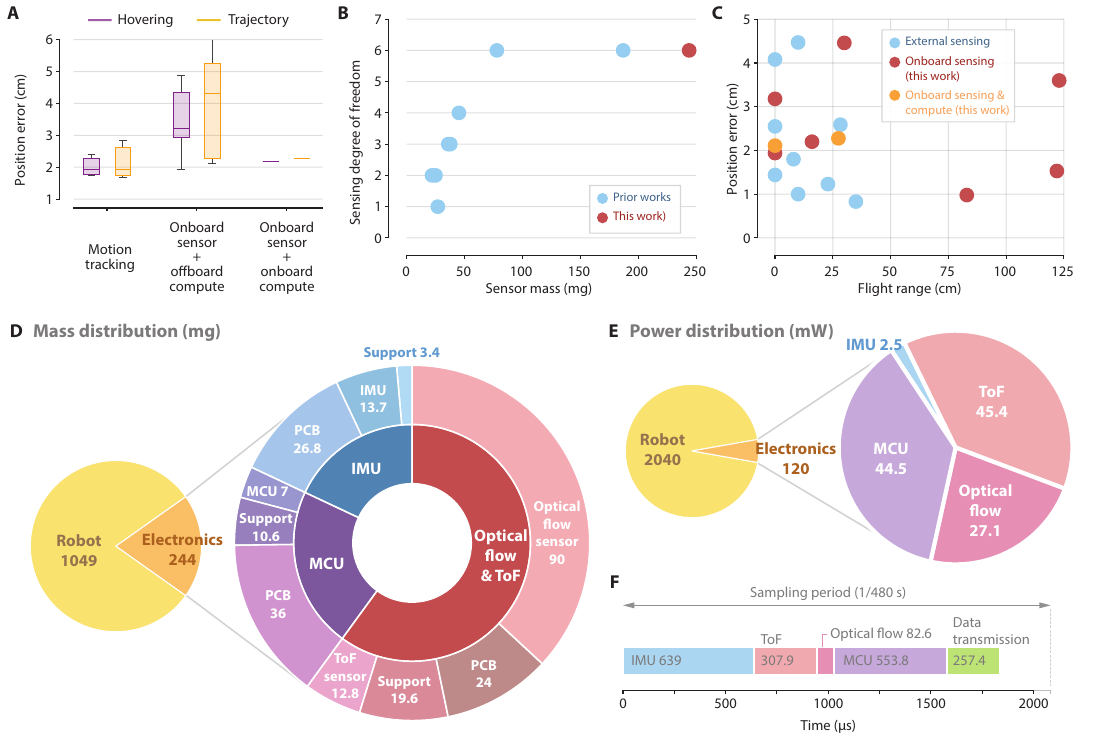}
\vspace{-8mm}
\caption{\textbf{Robot performance comparison and future design opportunities.} A) Comparison of robot flight accuracy between onboard and offboard sensing and computation. B) Number of estimated robot states as functions of sensing package mass. C) Comparison of robot flight accuracy and characteristic flight path dimension. D) Overall design and flight package mass distribution. E) Overall design and flight package power consumption distribution. F) Computation sequence analysis shows MCU only consumed less than 25\% of computational resources. }\label{fig:sensor-compare}
\end{figure}
In this work, we developed a 244-mg flight electronics package that enables onboard sensing and computation for a 1.29-g bio-inspired flapping-wing robot. By integrating a state estimator with a high-bandwidth low-level feedback controller, the robot successfully executed a range of flight maneuvers, including hovering, setpoint switching, and trajectory tracking. Across these experiments, the robot achieved positional accuracies of 2.21 – 6.1 cm over time horizons of 6 – 12 s, demonstrating its potential to perform insect-like maneuvers in complex and constrained environments. We further implemented a hierarchical control architecture that allows a human operator to issue real-time high-level commands. Using this framework, the robot completed multiple 15–30 s flights in cluttered environments, during which it navigated around obstacles and successfully landed on a sunflower. Together, these results represent a significant milestone for the IMAV community, which has pursued sensory-autonomous flight for more than a decade.

Our key insight was to adopt a task-driven design methodology that informed both robot redesign and sensor selection. Achieving stable landing on a delicate flower requires the robot to be lightweight while maintaining centimeter-level flight precision. These requirements led us to systematically evaluate the tradeoffs between mass constraints, payload capacity, and flight performance. We evaluated a range of commercially available micro-electromechanical systems (MEMS) sensors and selected the lightest sensor configuration that met the required sensing frequency and accuracy. During this process, we also determined that an appropriate PCB thickness was necessary to reduce vibration-induced noise and sensor drift. Fig. \ref{fig:sensor-compare}A compares robot flight accuracy under three conditions: offboard motion capture and computation, onboard sensing and offboard computation, and onboard sensing and computation. We found that switching to onboard sensors led to a 65\% increase of position error for hovering, while implementing onboard computation had a much smaller effect. Our robot maintained centimeter-scale flight accuracy - critical for performing challenging tasks such as obstacle avoidance and landing on a flower.   

Our flight package was the first system that enables sensory and computation autonomous flight in IMAVs. Despite weighing more than other sensor packages, our system can estimate full state information including attitude, angular velocity, position, and translational velocity (Fig. \ref{fig:sensor-compare}B). Compared to the other two systems that could estimate full robot states \cite{talwekar2022towards,yu2025tinysense}, our flight package achieved substantially lower attitude and position error over longer time durations, and it was evaluated in real flight conditions without requiring external motion capture. This result allowed the robot to fly outside of a motion capture environment (Fig. \ref{fig:operator}) and extend its operation range. Fig. \ref{fig:sensor-compare}C compares robot flight trajectory dimension and the associated positional accuracy. Compared to other IMAV platforms that require offboard motion capture, our robot achieved similar flight accuracy with only onboard sensors and MCU.

Integrating the flight electronics onboard required the robot to accommodate additional payload mass. By scaling the soft artificial flight muscles, we achieved a 46\% increase in net lift force. This redesign process highlights volumetric scalability as a key advantage of dielectric elastomer actuators (DEAs) over alternative microscale flight actuators. The DEA output force, displacement, and resonance frequency can be tuned by scaling the actuator radius and length. Importantly, DEAs can be scaled volumetrically while maintaining an approximately constant output power density. In contrast, rigid capacitive actuators, such as piezoelectric bimorphs \cite{woods2007energy}, have a fixed material thickness. As vehicle size increases, the required planar actuator area grows disproportionately, substantially complicating robot redesign. While recent studies \cite{yu2025tinysense,naveen2025hardware} attempted to adapt flight electronics to existing IMAV platforms, mismatches between sensory requirements and available payload capacity precluded fully onboard autonomous flights. In contrast, through the co-design of flight electronics and the robot, we achieved flights with onboard sensing and computation. The DEA volumetric scalability will further facilitate future robot redesigns aimed at integrating onboard power electronics and energy sources.

The co-design of the robot and flight electronics provided insights into IMAV weight distribution and power budget (Fig. \ref{fig:sensor-compare}D-E). At the gram scale, overall system mass and power consumption were dominated by the flight actuators, which accounted for 81\% of the total system weight and 94\% of total power consumption, respectively. In contrast, the flight electronics constituted a smaller share, contributing only 19\% of the system weight and 6\% of the overall power consumption. From a power budget perspective, there are opportunities to substantially reduce the MCU power consumption. According to the MCU datasheet, the integration of a lightweight ($<$5 mg) buck regulator is expected to reduce power consumption by approximately 50\%. In addition, computation sequence analysis (Fig. \ref{fig:sensor-compare}F) shows the low-level controller only utilizes only 25\% of a single core’s computational capacity. This under-utilization suggests that additional power saving could be achieved by reducing the MCU clock frequency and disabling the unused core. 

Beyond reducing MCU power consumption, a higher-impact direction is to leverage the available computational capacity to implement a low-rate, high-level planner. In the longer term, additional ToF sensors and a compact camera \cite{ozturk2024vision} can be integrated for the purpose of obstacle detection and navigation. ToF sensors positioned along the robot’s lateral axes would enable obstacle detection and avoidance, while a small camera could support object identification and high-level trajectory planning. Based on the landing demonstrations conducted by a human operator, we estimate that the high-level trajectory planner would require an update rate of only 1 – 2 Hz. Accordingly, we believe it is feasible to implement a planning algorithm that can operate in real time on the same MCU. Finally, the co-design framework can be extended to investigate power autonomy. We estimate that integrating power electronics and onboard batteries will require an additional payload of approximately 1 – 1.5 g, necessitating a 15–35\% increase of robot size relative to the platform presented in this work. We anticipate that continued advances along these directions will enable fully autonomous IMAV flights in the near future.


\clearpage 

%
\bibliography{science-bib} 
\bibliographystyle{sciencemag}


\subsection*{Acknowledgments}

\noindent\textbf{Funding:} \\
\hspace*{2em} National Science Foundation FRR-2202477 (YC) \\
\hspace*{2em} National Science Foundation FRR-2236708 (YC) \\
\hspace*{2em} Office of Naval Research N00014-25-1-2303 (YC) \\
\hspace*{2em} Air Force Office of Scientific Research MURI FA9550-19-1-0386 (JPH) \\ 
\hspace*{2em} MathWorks Engineering Fellowship (YHH) \\
\hspace*{2em} MathWorks Research Equipment Grant (YC) \\
\hspace*{2em} Zakhartchenko Fellowship (YHH, SK) \\

\noindent\textbf{Author contributions:} \\
\hspace*{2em} Conceptualization: YHH, QPK, ZG, JPH, EFH, YC \\
\hspace*{2em} Methodology: YHH, QPK, ZG, JC, YC \\
\hspace*{2em} Software: YHH, QPK, ZG, JC  \\
\hspace*{2em} Validation: YHH, QPK, ZG, SK, JC, OM  \\
\hspace*{2em} Formal analysis: YHH, QPK, YC \\
\hspace*{2em} Investigation:  YHH, QPK, ZG, YC \\
\hspace*{2em} Visualization:  YHH, QPK, ZG, JC \\
\hspace*{2em} Funding acquisition: JPH, EFH, YC \\
\hspace*{2em} Project administration: JPH, EFH, YC \\ 
\hspace*{2em} Supervision: JPH, EFH, YC \\
\hspace*{2em} Writing – original draft: YHH, QPK, YC \\
\hspace*{2em} Writing – review \& editing: YHH, QPK, ZG, SK, JC, OM, JPH, EFH, YC 

\paragraph*{Competing interests:}
The authors declare no competing interests.

\paragraph*{Data and materials availability:}


\subsection*{Supplementary materials}

\hspace*{2em} Supplementary Text\\
\hspace*{2em} Figs. S1 to S25\\
\hspace*{2em} Table S1\\
\hspace*{2em} Movies S1 to S7\\


\end{document}